\documentclass[twocolumn,10pt]{asme2e}

\usepackage{graphicx}
\graphicspath{ {images/} }

\confshortname{DSCC 2017}
\conffullname{the ASME 2017 Dynamic Systems and Control Conference}

\confdate{October 11-13}
\confyear{2017}
\confcity{Tysons Corner, Virginia}
\confcountry{USA}

\papernum{DSCC2017/5177}

\title{Development of a Swarm UAV Simulator Integrating Realistic Motion Control Models For Disaster Operations}

\author{Kazi Tanvir Ahmed Siddiqui and David Feil-Seifer
    \affiliation{
Robotics Research Laboratory\\
Computer Science \& Engineering Department \\
University of Nevada, Reno\\
Reno, Nevada, 89557\\
Email: kahmedsiddiqui@nevada.unr.edu | dave@cse.unr.edu

    }	
}

\author{Tianyi Jiang, Sonu Jose, Siming Liu, Sushil Louis     
    \affiliation{Evolutionary Computing Systems Lab\\
Computer Science \& Engineering Department \\
University of Nevada, Reno\\
Reno, Nevada, 89557\\
    }
}

\begin{document}
\maketitle    

%%%%%%%%%%%%%%%%%%%%%%%%%    Abstract Text   

\begin{abstract}
{\it Simulation environments for Unmanned Aerial Vehicles (UAVs) can be very useful for prototyping user interfaces and training personnel that will operate UAVs in the real world. The realistic operation of such simulations will only enhance the value of such training. In this paper, we present the integration of a model-based waypoint navigation controller into the Reno Rescue Simulator for the purposes of providing a more realistic user interface in simulated environments. We also present potential uses for such simulations, even for real-world operation of UAVs.
}
\end{abstract}

%%%%%%%%%%%%%%%%%%%%%%%%%%%%%

\section{Introduction}

Unmanned Aerial Vehicles (UAVs) are being utilized in civilian airspace for emergency operations. These applications include providing overwatch for search and rescue and helping emergency management personnel define the boundaries of forest fires. The critical value of such systems is to increase the situational awareness of emergency management personnel in disaster settings. Given reasonable concerns regarding the ability of an autonomous UAV to properly operate in such a setting, semi-autonomous systems where a human operator is involved in decision-making but does not exert total control are preferred. An operator would provide a UAV or a group of UAVs with high-level tasking and the UAV would carry out such a command.

% uses for simulators for training/user interface prototyping
The conditions under which such public safety UAV swarms would operate do not occur regularly and have little room for error. As such, there is a need for simulation environments to be used for interface prototyping and operator training. To support these goals, we have developed a UAV swarm disaster simulator. We use this simulator to study human operators' effectiveness in long-term search and rescue operation of a swarm UAV in a simulated earthquake setting. Through the use of a controlled environment such as this, we are able to evaluate interface design choices with regard to operator situational awareness.

% study of operator factors for long-term UAV operation
This paper details the development of a simulator utilizing a video game-like interface (based on real-time strategy games). An initial validation where UAV flight was very idealized will be described. The motion controller did not include any physical properties of the UAV (other than simple linear/angular velocity limits) for its motion planning. We then integrate an established aerodynamic model of UAV flight into our existing simulator. We present some preliminary validation of this system, and hypothesize on some ways that interfaces might utilize model-based simulations of flight trajectories, even for real-world operation of a UAV.

The remainder of this paper is organized as follows: Section II presents background and some related work, followed by a general system architecture in Section III. Section IV details the experimental design. We present the results in Section V. Section VI describes the simulator redesign for more accurate model-driven trajectory behavior. Finally in Section VI, discussion is presented followed by conclusion and future work in Section VII.

\section{Background}

\begin{figure}[tp]
	\centerline{\includegraphics[width=3in]{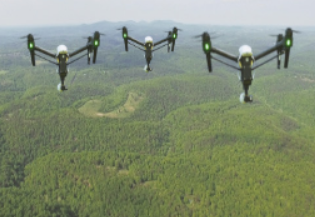}}
	\caption{Unmanned Aerial Vehicles (UAVs) flying over forest environment. The field of view of one UAV equipped with a camera is very small, hence multiple UAVs would be used for emergency management scenarios.\label{fig:uav}}
\end{figure}

Using commercially-available software, flight simulations can be used for pilot training and simulating scenarios both typical and atypical~\cite{bruce2006microsoft}. Recent advances in computing power, especially in graphics, and in software environments for 3D modeling and simulation enable scientists and engineers to quickly prototype and use immersive, interactive, virtual models of real-world scenes. In addition, the growth in cheap computing power has advanced Artificial Intelligence (AI) and consequently autonomy in cyber systems, e.g. the current popularity of self-driving car engineering with global automakers~\cite{guizzo2011google}. These advances are crucial enablers for prototyping and evaluating simulation systems and for connecting such systems to autonomous robots operating in the real world. 

There are many tools to simulate UAV operations. Simulations not only offer a platform for faster and safer operations but also enable reduced resource costs for operator training. USARSim~\cite{carpin2007usarsim} is one of the applications that is built on the top of Unity Engine to simulate multiple robots and a flight environment. The Mobility Open Architecture Simulation (MOAST)~\cite{goncalves2014game} provides an interface for robot control and customization. Users can modify the existing modules or they can add more modules to create more complex robots than in USARSim.

Neptus~\cite{dias2006mission} utilizes a modular approach consisting of the following components: The mission console to interface with the other modules as well as for mission execution; the mission planner, responsible for mission setup which includes map generation and user friendly interface; the mission reviewer, mainly for collecting information during mission execution and also performs the analysis of past mission’s data; the mission data broker, handles data and for accessing those data it offers web services; and finally vehicle simulator for simulating real vehicles concurrently~\cite{goncalves2014game}.
	 	 	
\begin{figure}[htp]
	\centerline{\includegraphics[width=3in]{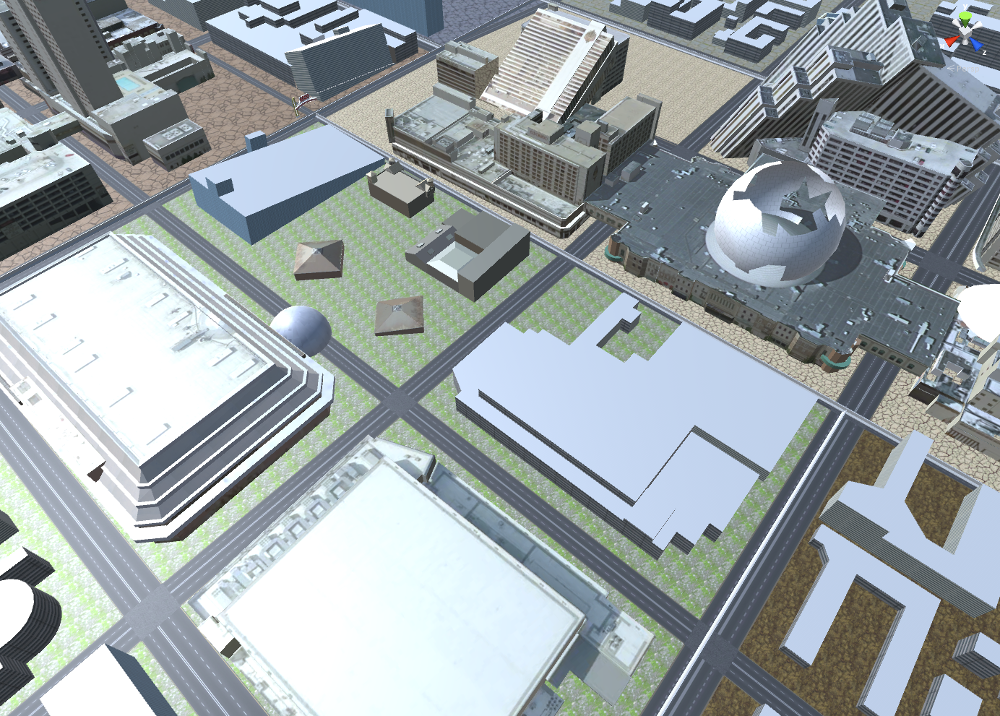}}
	\caption{Simulation rendering of a damaged Reno downtown area in the Reno Rescue Simulator. This matches the actual layout of the city of Reno,  but some buildings have been damaged and collapsed.  \label{fig:reno}}
\end{figure}

A review of mobile robot simulation environments reveals that simulation is becoming an increasingly important aspect of mobile robots~\cite{josh2006video}, helping researchers perform more experimentation in this area. A realistic graphical rendering system and ideal physics simulations are the main features of a best simulator. Computer video games engines often are used to power a robot simulation environment, capable of simulating multiple robots, people, and objects in the environment.

A more recent paper proposed a UAV-based solution to help on the search and rescue activities in disaster scenarios~\cite{camara2014cavalry}. These UAVs are specialized to perform operational tasks (e.g., providing a temporary communication structure, creating up-to-date maps of the affected region and searching for hot spots where the rescue teams may have more chances of finding victims) and attain search-and-rescue objectives. These robots utilize sensors fixed on the UAVs, such as infrared cameras, radars, or portable devices for detecting radio signal~\cite{camara2014cavalry}. All of these activities require specific competences, and as such, more than one UAV or sensor type may be required to accomplish all of them. This UAV-based fleet, to be efficient and useful in the terrain needs to be semi-autonomous and more capable of self-organization. 

%DFS->Tanvir: need a paragraph, here about the model developed for your trajectory planning

Michael et. al. \cite{michael2010grasp} developed a mathematical model for planning the trajectory of a UAV. To move from point A to point B,  the error in $x, y, z$ position and used the error to determine $\delta_{a_x}$, $\delta_{a_y}$, and $\delta_{a_z}$, the required changes in acceleration. To calculate the angular speed for each of the rotors and their orientation, the changes in rotation angles $e_\theta$, $e_\phi$, and $e_\psi$ are calculated. Using these values, the angular speed of the motors: $\Omega_1$, $\Omega_2$, $\Omega_3$, and $\Omega_4$ can be determined. Using the values of angular speed of the rotors, we calculate the propeller force, moments, and inputs for the rotors. %The new input changes the position and orientation of the UAV. After time $dt$ the whole process is repeated and thus the UAV keeps flying.
%(that's enough)

\section{Simulator Development}

Computer-based simulations have been used in operational training and research for complex systems like airplanes, robots, and military equipment. In this section, the design influences for the system presented in this paper are outlined.

In emergency management scenarios, a UAV operator would likely need to control UAVs while following directives from a scene commander and communicating with a team in the field. This combination of skills required for effective operations likely means that an operator would need significant practice in order to effectively function during a disaster. Since earthquakes and wildfires are not predictable, such training could more regularly be delivered through a simulation.

We developed our simulation system $RenoRescueSim$ based on the Unity3d game engine. This simulation allows an operator to control multiple UAVs for cooperative search in a large-scale urban area modeled after a real city (Reno). An operator can observe the world either through a top-down ``map'' view (see Figure~\ref{fig:gui}) or through two ``first-person" views (one facing forward on the UAV, and one facing downward; see Figure~\ref{fig:reno}). The map view can be used to localize where the UAVs are in the simulated environment, and the first-person views can be used to observe where people, damaged vehicles, or fires are as they search through the city.

Our system also provides an interface for search route planning so that the operator can plan several actions simultaneously to accurately search and promptly rescue people (see Figure~\ref{fig:uavnav1}). The simulation can be used either for operator training to enhance user's proficiency with operating multiple UAVs or to evaluate the performance of the system or to test features of user interface and their effect on the situational awareness, frustration, and physical and mental demands utilizing the system places on an operator. 

\begin{figure}[t]
	\centerline{\includegraphics[width=3in]{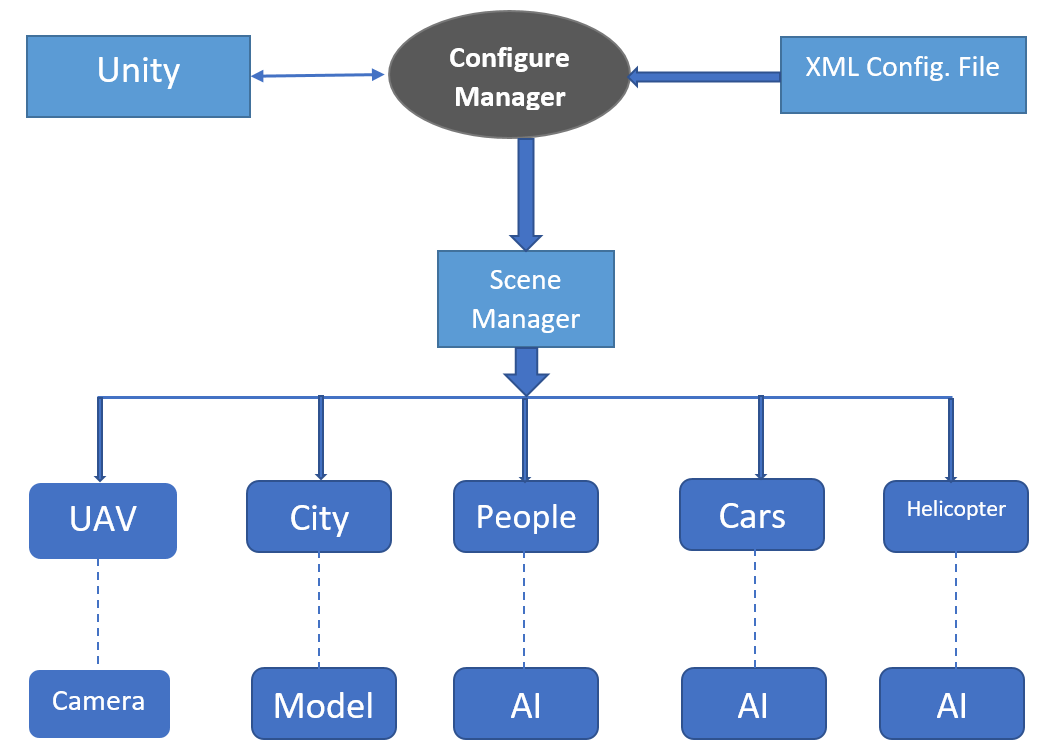}}
	\caption{System Architecture. The simulator developed in Unity game engine. The configure manager can be changed using XML configuration file using Unity. The Configuration manager changes Scene manager. The scene consists of UAV, which has a camera. The camera is used to render view from the UAV. The city is modeled, except the map , which is rendered from Google maps. There are numbers of random people, cars and helicopters roaming in the city. Their movements are controlled using AI }
	{\label{fig:arch}}
\end{figure}

\subsection{System Architecture}

Current challenges for developing an effective simulation include accurate physical and behavioral characterization of a robot along with its sensors, effectors, and instruments; seamlessly networking real and virtual worlds; properly trading-off manual and autonomous control for optimal task performance; and building models that predict how human performance in simulation environments transfers to performance in the real-world. What is clear, however, is that it is possible to train for scenarios in simulation that would be very difficult and expensive in reality. Without this training, especially for recovering from error states, operators may inadvertently lose valuable hardware, produce erroneous results, and compromise system and human safety~\cite{law1991simulation}.

Figure~\ref{fig:arch} shows the system architecture of our Search and Rescue Simulation. Our system consists of two major components including the \textit{Configure Manager} and the \textit{Scene Manager}. The functioning of the UAVs, distribution of objects throughout the city, and the movements of people, cars and helicopters are handled by the scene manager.

\subsection{3D Environment Modeling}\label{sec:env}

Game Engines Simulations form a core component of many computer games. Game engines such as Unity3d, Cryengine, and Unreal all provide commercial strength and well-supported 3D game development environments that enable fast, reliable, 3D simulation development~\cite{mat2014using}. Unity3d in particular has distinguished itself as a popular multi-platform tool for training simulation development~\cite{blackman2013beginning}.

\begin{figure}[ht]
	\centerline{\includegraphics[width=3in]{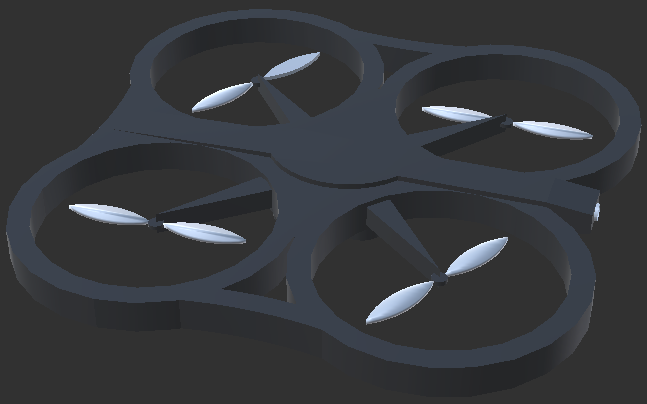}}
	\caption{UAV model used in the simulator. This UAV has two cameras. One for looking directly in front and other for looking below. It also simulates all the sensors and actuators the hummingbird robot has (e.g., GPS, wireless communication, and inertial guidance system). \label{fig:quad}}
\end{figure}

\begin{figure*}[th]
	\centerline{\includegraphics[height=3in]{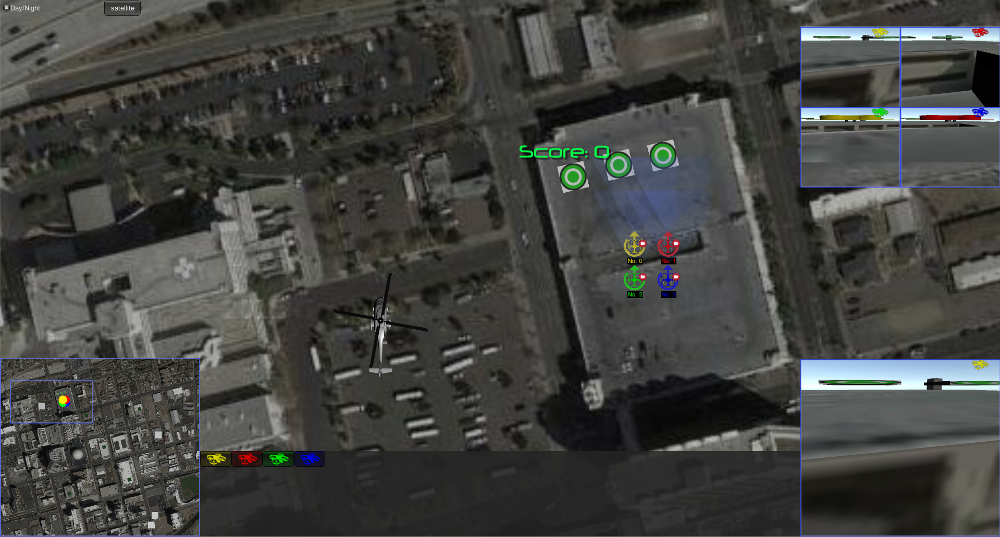}}
	\caption{$RenoRescueSim$ user interface. {\bf Center}: is the main view, where a user can get a top-down view of the world (rendered from Google Maps tiles); {\bf Bottom-Left}: minimap view of the entire city, with area viewable in the center panel shown (blue-box); {\bf Bottom-Right}: View from currently-selected UAV's camera; {\bf Top-Right}: Views from all UAV's cameras}
	{\label{fig:gui}}
\end{figure*}

In addition to the availability of 3D game engines for simulation development, 3D models of people, vehicles, buildings, and many common objects are now more easily available in web stores (e.g. Unity Asset Store or TurboSquid). The assets available and their support for data available from mapping services like Google Earth~\cite{patterson2007google} make it possible to quickly model a specific city, specific unmanned aerial vehicles, and specific scenarios in a real-world task simulation. Operating and training through rehearsal in a high-fidelity simulation environment has the potential to translate to better, faster operation in the real world~\cite{nicolescu2007training}. However, developing familiarity with robot capabilities, instruments, instrument readings, and behaviors will also require investigating and building non site-specific training and operator assessment tools.

In order to improve the realistic level of simulation system, our 3D virtual environment is modeled after the urban area of a real city (in this case, Reno, NV). Main buildings and streets of city Reno are modeled and textured in our 3D virtual environment. The UAV model we created (as shown in Figure~\ref{fig:quad}) contains propeller, body frames, motors to define the actuation of the system as similarly as possible to the model presented in~\cite{michael2010grasp}. Our UAV model can be easily adapted into the validation of physical control if needed. We have also equipped the UAV with simulated sensors, such as GPS and cameras (forward and downward-looking) to simulate the ways that information could be relayed from the UAV to an operator.

\subsection{Scene Manager}

The Scene Manager connects the simulation of the real world environment with the Unity3d engine. The Configuration Manager provides the ability to parameterize the Unity simulation with an XML configuration file. These configuration variables are used to coordinate between the game application which is the interface used to interact with the system, the simulated people and cars and the game missions. With the XML configuration file, we can quickly develop controlled experiments examining elements of the system (UI elements, autonomy levels, etc.) and analyze user interaction data collected during usage of the system for the purposes of post-hoc analysis. The behavior of injured people, cars, fires, etc. are simulated using simple heuristic behaviors.

\subsection{User-Interface Design}

Figure~\ref{fig:gui} shows the graphic user interface of Reno Rescue simulation during search and rescue mission. The snapshot shows initial scene of our simulator. Four UAVs are placed initially at a particular location, available for an operator to command their navigation to other locations. The top-right window shows the camera view of all UAVs whereas bottom-right represents camera view of a single UAV. When users double click on a UAV or when they double click on the top right camera view of any UAV, then they can see the camera view of that particular UAV in a  rectangular window in the bottom right of the scene.

\begin{figure}[hbp]
	\centerline{\includegraphics[width=3.0in]{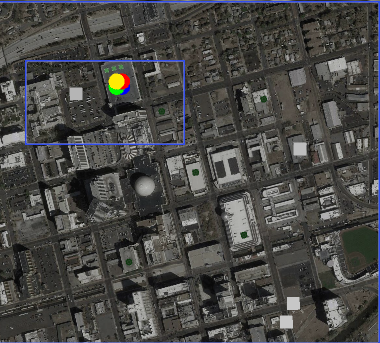}}
	\caption{Minimap panel used to show entire map with locations of all UAVs (red, yellow, green, and blue dots). Current view of center panel is shown (blue rectangle). Users can change the center panel view or set waypoints from this panel.}
	{\label{fig:minimap}}
\end{figure}

The center of the screen is a perpendicular camera covers a small area of the city. This area is a main operating interface for user to control multiple UAVs and design the navigation search path. The simulation also provides a bird view camera for the entire search area on the bottom left of the screen to show the overall status of the search task, as shown in Figure~\ref{fig:minimap}. Note that both the center area and the mini-map are covered by satellite view downloaded from Google Map. The UAV Control area locates at the bottom of the screen. User can select a single UAV by click a UAV icon in the UAV Control Area. The bottom right area shows a camera view of the selected UAV. The camera shows the real world 3D model environment described in Section~\ref{sec:env}.

\begin{figure}[htp]
	\centerline{\includegraphics[width=3.6in]{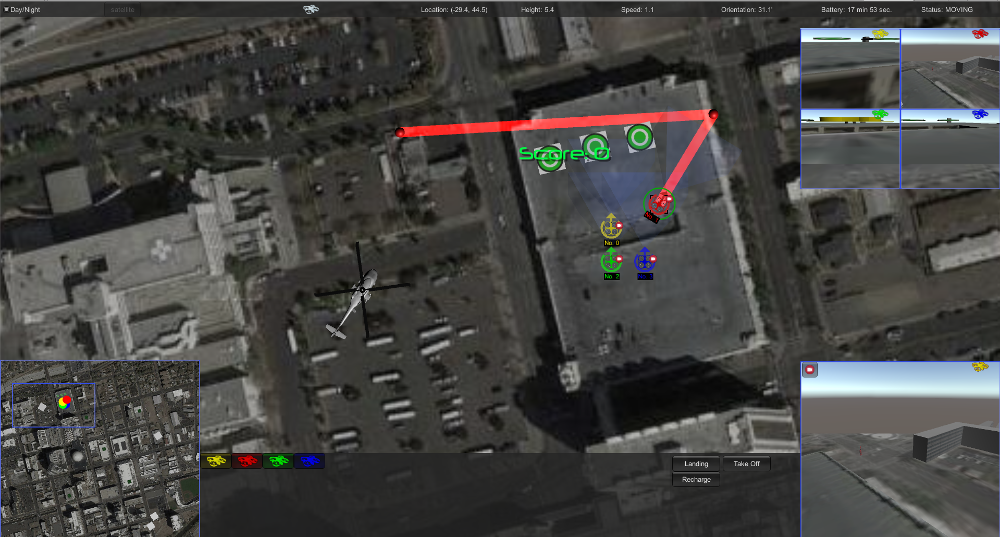}}
	\caption{Single-UAV navigation with two sequential waypoints set (red lines) (Level 2, 3 autonomy cases). Waypoints can be set by selecting a UAV and right-clicking either in the center panel or in the minimap view.}
	{\label{fig:uavnav1}}
\end{figure}

\begin{figure}[htp]
	\centerline{\includegraphics[width=3.6in]{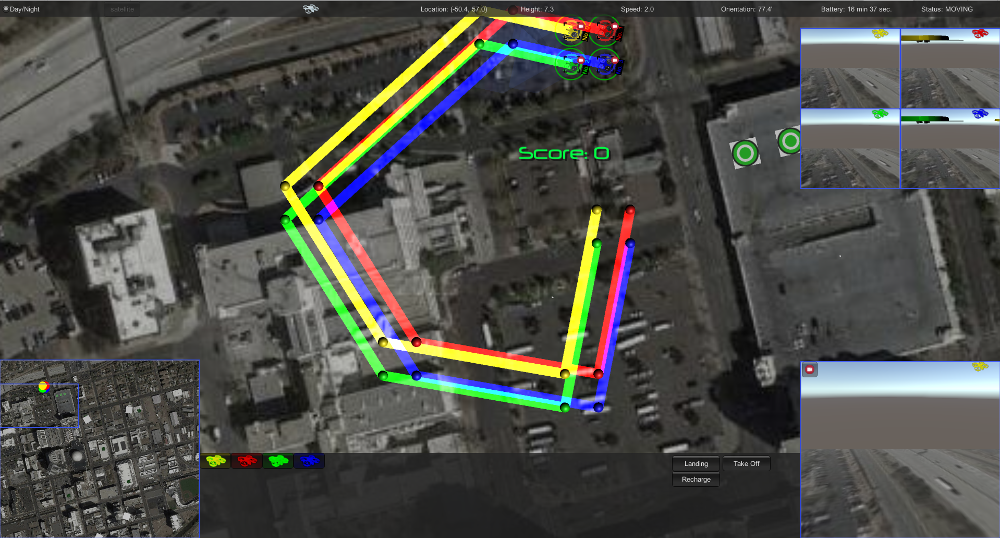}}
	\caption{Multiple-UAV navigation with several waypoints set for multiple UAVs to move in formation (Red, Yellow, Cyan, Blue lines) (Level 3 autonomy).Waypoints can be set by selecting multiple UAVs and right-clicking either in the center panel or in the minimap view.}
	{\label{fig:uavnav4}}
\end{figure}

Figure~\ref{fig:uavnav1} shows the navigation of a single UAV. It gathers the data at it moves. It can move to any location depending on the user activity. It can also increase or decrease its speed as well as pause option to stop its movement further. Figure~\ref{fig:uavnav4} shows the movement of multiple UAVs. All selected UAVs can start at the same time and move together to a particular location. It is also possible to view from all UAVs' cameras at the same time.

\section{Experimental Validation}

Our initial validation of this system addresses two research questions. First, can a RTS-style interface be effective for operator control of multiple UAVs for a simulated search-and-rescue task? Second, does prior experience with RTS-style games provide an advantage for operators training with the simulator?

To answer these research questions, we developed a user interface for our simulator that follows conventions from the popular commercial RTS game StarCraft II. We then designed UAVs with three different levels of autonomy for controlling and conducted experiments on evaluating the performance of conducting search and rescue tasks. And to provide operators with an adjustable scale of task difficulty to test their usage of the interface. The three levels of autonomy UAVs are described as follows:

\begin{itemize}
    \item \textbf{Level 1}: an operator can only control one UAV through direct flight control (increase/decrease altitude, move forward, backward, turn in place, slew left/right).
    \item \textbf{Level 2}: an operator is able to control one or more UAVs either through direct controls or by setting a single destination waypoint.
    \item \textbf{Level 3}: an operator is able to set one more more waypoints for a UAV to follow in sequence.
\end{itemize}

We performed a 3x3 between- and within-subjects study with two factors: autonomy type and trial number. Autonomy type has three levels (described above). Trial number has three levels: one, two, and three. We examined the simulator behavior using several dependent variables: situational awareness, mental demand using the simulator, physical demand, and frustration.

To measure situational awareness, we asked users after each 5-minute trial to answer questions related to the health and location of their UAV fleet. We asked users to estimate the battery level left (the level starts at 100, and decreases based on the amount of movement and time in the air, which can be regenerated by navigating back to the ``home base'' for the UAVs). The actual battery level is compared to the estimated level to get an accuracy measure.

We used the NASA Task-Load Inventory (TLX)~\cite{hart2006nasa} to estimate a user's mental and physical demand as well as their frustration with the interface after each trial. This is a well-established scale to measure an operator's effort when completing tasks, and has been applied for many general problems, especially user interfaces.

We hypothesized the following:

\begin{itemize}
    
    %\item {\bf H2}: Users with RTS experience will perform the task better than users without RTS game experience. This can be measured by comparing the task performance (numbers of cars or people found during the search-and-rescue task) between RTS and non-RTS users.
    
    \item {\bf H1}: Proficiency in the search-and-rescue task will increase the more an operator uses the simulator (practice effects). This can be measured by comparing operator situational awareness changes over time (higher is better) and by comparing the mental and physical demand of using the simulator (lower is better).
    
    \item {\bf H2}: The robot autonomy type (described above) will affect user demand and frustration with the interface (lower is better). The users will perform better with greater autonomy.
\end{itemize}

\subsection{Participant Recruitment}

Fifteen undergraduate and graduate students (10 Males, 5 females) with no prior experience with rescue operations or simulations of UAVs were recruited from the department of Computer Science and Engineering, University of Nevada, Reno. We recruited them by sending an email to invite them to participate in the experiment. Interested participants signed up for a 45 minute time slot. The participants' age ranged from 17 to 25 years. All are regular users of computers. Most of them were familiar with playing on-line computer video games.

\subsection{Experiment Procedure}

After welcoming the participants the experimenter gave some basic information about the purpose of this study asked them to sign a consent form. Prior to the experiment, participants completed a pre-test questionnaire soliciting demographic data, computer expertise, and familiarity with video games. The experiment began with a training session to acclimate the users to the simulator and its operation and the search and rescue goals. The experimenter demonstrated how to move the UAV to various locations and also how to play the game by locating the people and the cars and the scoring system. The training session was followed by actual experiment. The NASA Task Load Index (TLX) as well as a situational awareness questionnaire (see Appendix~\ref{app:sitawareness}) are presented to the operator 3 times, once after each 5-minute trial assessing the operator's awareness of the scene and the UAVs they are controlling. Participants were asked to accomplish the game tasks quickly and efficiently. 

\begin{figure}[t]
\centerline{\includegraphics[scale=0.8]{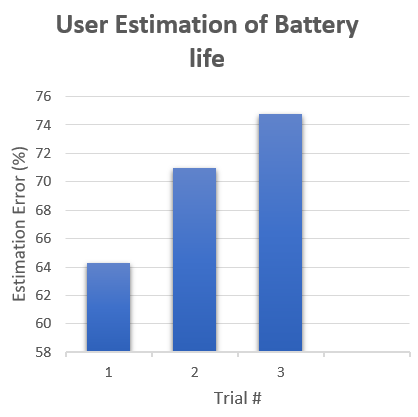}}
\caption{Situational awareness (higher is better) of a user by time spent interacting with the simulator expressed by the user's accuracy at estimating the remaining Battery Life of the UAV fleet.\label{fig:batt}}
\end{figure}

\subsection{Experiment Setup}

The experiment was performed in the ECSL (Evolutionary Computing Systems Lab), University of Nevada Reno. It is a quiet room with no background noise so that participants were able to concentrate more on the game. The participants were asked to use a Windows workstation running the simulator. The computer used an Intel Core i5 processor and 16 GB RAM. The system would execute the simulator and collect data from the experiment. We collected data from each trial using the data logger built into the simulator for storage in XML files. These data included responses to the questionnaires, and in-simulation usage data (actions-per-minute, overall health of the UAV swarm)

\subsection{Results}

To examine hypothesis {\bf H1}, we compared the values of situational awareness, mental demand, physical demand, and frustration for each time trial. For {\bf H1} to be supported, situational awareness will increase and the others will decrease as more time is spent with the simulator.

Figure~\ref{fig:batt} shows the operators' accuracy estimating the battery life of UAVs during each experiment trial. The data show that as the operator gains more experience with the simulator interface, the users' accuracy estimating UAV battery life improved. This accuracy increase suggests that operator situational awareness increased with simulator practice. While these results were not significant, it is likely that a larger sample size will improve the significance of these results.

\begin{figure}[t]
\centerline{\includegraphics[scale=0.65]{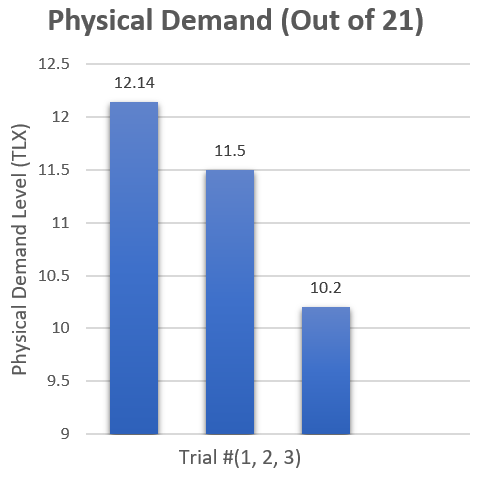}}
\caption{Physical demand (lower is better) reported by operators of the UAV after each trial (NASA-TLX survey). The decreased physical demand with each successive simulation trial indicate that training with the simulator makes it easier to use ($p < 0.001$). \label{fig:physdemand}}
\end{figure}

\begin{figure}[t]
\includegraphics[scale=0.65]{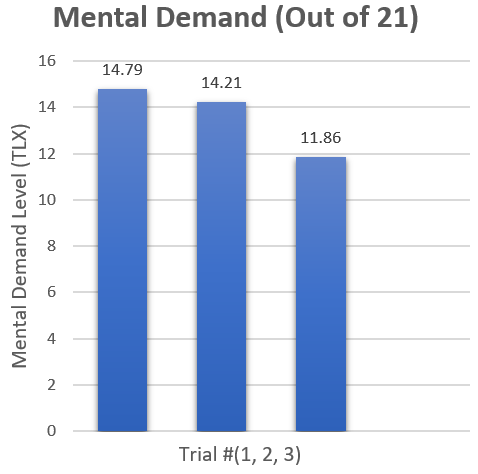}
\caption{Mental demand (lower is better) reported by operators of the UAV after each trial (NASA-TLX survey). The decreased physical demand with each successive simulation trial indicate that training with the simulator makes it easier to use ($p = 0.058$). \label{fig:mental}}
\end{figure}

Figures~\ref{fig:physdemand} and~\ref{fig:mental} show the users' mental and physical demand level (measured by the TLX survey) by trial. Later rounds show less mental and physical demand was required (differences were not significant). This demonstrates that the more experience a user has with the simulator interface, their cognitive load decreases, demonstrating a training effect. %need frustration values, here

To examine hypothesis {\bf H2}, we compare the same factors with autonomy level as the independent variable. We conducted a MANOVA with Physical Demand, Mental Demand, and Frustration as the dependent variables and autonomy level as the independent variable. The multivariate result was significant for autonomy level, Pillai's Trace = 0.43, $F = 4.46$, df = 36, $p < 0.01$. Follow-up univariate tests showed that Frustration was significant, $p < 0.001$ and Mental Demand was marginally significant, $p = 0.058$. Tukey's HSD tests showed that Levels 2 and 3 were significantly lower than Level 1.

% RTS/non-RTS results (speak with T before including)

%\begin{figure}[h]
%\includegraphics[scale=0.9]{Capture10}
%\caption{RTS user Vs Non RTS user\label{fig:rts-nonrts}}
%\end{figure}

%Figure~\ref{fig:rts-nonrts} shows a comparison of RTS and Non RTS user performance during the game. As you can see from above, the mean of finding cars by RTS game users is 45.5 which is higher than non RTS game users whose mean is 25.3. In our scene we have a total of 60 units( 30 cars, 30 people). Hence our hypothesis is validated by showing the result that RTS game users perform
%80\% more better than non RTS game users.

\subsection{Discussion}

These data partially support hypothesis {\bf H1}, showing that there is a trend in the direction pointed to by the hypothesis, but not enough to conclude that there is a training effect due to the simulator. It is likely that given more time, and a larger participant pool, the data would show a greater training effect.

These data support hypothesis {\bf H2}. Lower mental demand and frustration were observed when the robots behaved with more autonomy. These results make sense, since a user was able to more easily operate the UAVs while also performing the search-and-rescue task. It is likely that given a greater simulator time, the users would have as high or higher differences between the autonomy groups.

While these results are promising for the use of such as a system as a training simulator ({\bf H1}) and to evaluate elements of UAV user interfaces ({\bf H2}). As part of our collaboration with UAV researchers, we identified several areas where the UAV did not perform as accurately in simulation to what real-world behavior would be. As this could have significance on the training value of such a simulator, we wish to increase the realism, particularly of the UAV movement in simulation.

\section{Simulated Dynamics For More Realistic UAV Movement}

To address the realism of UAV movement, we turn to established models of UAV dynamics. Michael et. al. \cite{michael2010grasp} have provided an accurate aerodynamic model of micro UAV (MAV) flying. MAVs are between 0.1-0.5 meters in length, and 0.1 to 0.5 kilograms in mass \cite{kroo2000mesicopter}. MAVs are commonly utilized for civilian applications; therefore, they are the size class platform that we will simulate for this work.

\begin{figure*}[th]
	\centerline{\includegraphics[width=8in, height = 5in]{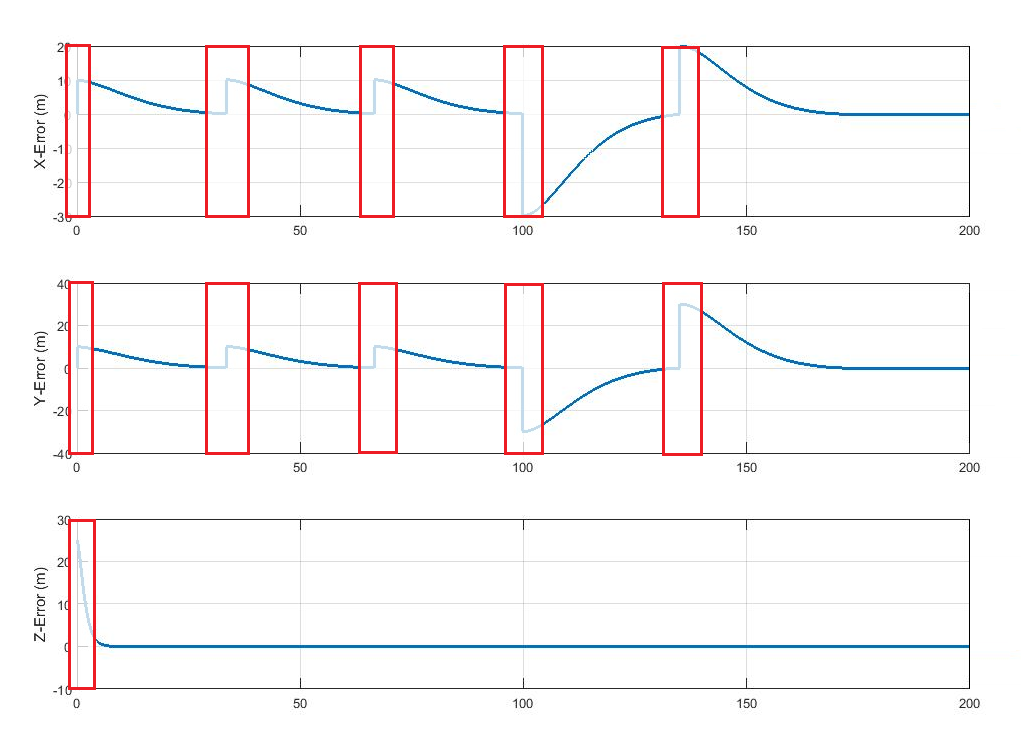}}
	\caption{Graph showing the difference in planned trajectory and current trajectory (waypoint transitions are indicated by red rectangles)}
	{\label{fig:poserror}}
\end{figure*}

Michael's model is specifically designed for the Hummingbird quadrotor sold by Ascending Technologies. It has a 55 cm tip-to-tip wingspan, 8 cm height, and 500 grams of weight including the battery. Also, it has a battery life of 20 minutes, and can carry 200 grams of payload \cite{michael2010grasp}. The small size and dexterity of Hummingbird UAV made it suitable to navigate through a constrained space. In this section, we implemented the aerodynamics of the Hummingbird UAV using the formulae by \cite{michael2010grasp}.

To make the UAV movement more realistic and suitable for training purposes, some critical aspects of UAV movement need to be considered. First, a UAV will not move from point A to point B at a uniform rate as the dynamics of the system need to be considered. Furthermore, as the UAV changes velocity in any direction, pitch and roll changes occur. As it is likely that a fixed camera on a UAV will be what an operator will use for a search-and-rescue task, simulating such attitude changes would be crucial for an operator's later proficiency with a real-world system.

We make our UAV's simulated flight path mimic the state of a UAV for real-world flight. Three axes $ x, y$ and $z$, that locate its position, and three angles $\phi$, $\theta$, and $\psi$ that measure the angular distance from respective axes are derived from this model. These variables control the movement and orientation of a UAV. Each of these variables are a function of the angular speed of the rotors. The angular speed of each rotor creates thrust and lift, which are opposed by the forces due to drag and gravity. Though the effect of wind is significant for the movement of such a small aircraft, we discarded the effect of wind in this simulation for the sake of simplicity. We assume that, the UAV will be flying in a closed environment where the effect of wind is nominal.

When an operator provides a series of points for the UAV to follow, the simulated controller will plan a trajectory to reach each goal, obeying the dynamics of the system. The UAV updates its flight path by using a Proportional Derivative (PD) controller. The movement and orientation of the UAV showed in the simulator represents real world UAV flying. This results in the simulator tilting while turning and pitching when accelerating/decelerating, which resembles a real world UAV flight. It also sets a more dynamically appropriate trajectory than a carrot-style planner. We wanted to simulate real world UAV flight for the purposes of training so that rescue operators would have a solid understanding of how such a system would move during emergency operations. 

We implemented the flight dynamics of the UAV in two steps. In the first step, we defined the physical properties related to the UAV flying. Those are: mass of the UAV, gravitational acceleration, thrust co-efficient for motor, distance from the center of the UAV to the rotors, PD control parameters (for controlling position and orientation), and moment coefficient for motors. 

Next, for each frame in the Unity game engine, we calculated the desired angular speed, rotational speed, attitude control parameters, force, moment, and inputs for each rotor, net force acted upon the UAV, and orientations (yaw, pitch, and roll angles) with respect to three axes. We compensated the error of the UAV from the desired flight path by using the PD parameter and added that error in every frame. Each frame in Unity represents the minuscule time interval $dt$. We used 60 frames per second for this simulation.

\begin{figure}[t]
\includegraphics[width=3.5in]{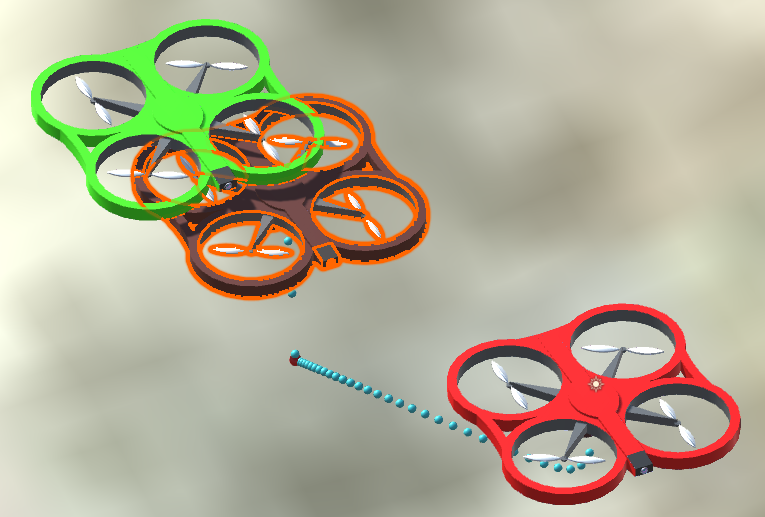}
\caption{Modified trajectory planner in action. The red UAV represents the start position; the brown UAV is the current position (trajectory shown with blue dots) and green UAV indicating the goal.\label{fig:3drones}}
\end{figure}

%DFS->Tanvir: need results section/paragraph

The simulation showed us how the UAV followed the flight path created by the autonomous flight dynamics algorithm. We show an example movement in Figure~\ref{fig:3drones}. Our simulator showed that the UAV (brown) followed a flight path and is able to repeatedly reached the destination. The flight path created by the UAV was reasonable and quick to implement. The flight path was direct and slowed its velocity when it was close to the destination (see Figure~\ref{fig:poserror}). Upon reaching the destination, the UAV hovered to maintain its position and orientation.

\section{Applications}

There is a clear possibility that more detailed simulations of UAV flight trajectories could lead to a better environment for training operators. However, having a realistic prediction for how a UAV might move given an operator waypoint command may have additional benefits beyond training.

% operator expression of error from idealized path

From a user-interface perspective, if the above model provides a planned trajectory for reaching a way-point, we can measure deviation from that trajectory and provide that for a user. This could be indicative of external factors (wind, jet-wash) or internal factors (damaged equipment, unbalanced UAV, sensor issues) which would be relevant for an operator to know. This could be rendered by showing predicted positions overlayed with actual positions (detected with IMU/GPS) or by showing error ellipses to indicate position error.

% detecting air disturbances (wind, etc)
Detecting consistent external factors could have additional uses. Wildfires can be modeled based on terrain, wind, and other weather conditions~\cite{cary2006comparison}. Furthermore, an effective method for UAV swarm control is to distributively and autonomously surround a wildfire to provide overwatch~\cite{pham2017distributed}. Such algorithms can use wind, measured from such error discussed above as added data for such predictive models.

% model for real-world behavior, simulate for future work

\section{Conclusions}

This paper presents preliminary development and validation of a multi-UAV simulator for disaster operations training. The initial validation indicates that there is potential value for use as a training simulator and that novel research in effectiveness of UAV autonomy on situational awareness can be accomplished using commonly-used metrics through this simulator. We also demonstrate model-based UAV movement for more realistic UAV trajectories that observe dynamics similar to a real-world UAV.

Future work will involve utilizing this UAV for longer-time operations to determine for what length of time training with the simulator produces added proficiency. Additionally, we will evaluate the added realism of the system with operator with prior UAV piloting experience to see if it more accurately reflects the flight profile for a UAV. We will then develop this simulator interface to act as an operator interface for a real-world UAV swarm.

\begin{acknowledgment}
The authors would like to acknowledge the financial support of this work by the National Science Foundation (NSF, \#IIS-1528137, \#IIP-1430328). We would like to acknowledge the help of Rumit Kumar and Manish Kumar.

\end{acknowledgment}

%\vspace{10pt}
%{\bf Keywords:} User Interfaces, Simulation, Search and Rescue.

%%%%%%%%%%%%%%%%%%%%%%%%%%%%%%%%%%%%%%%%%%%%%%%%%%%%%%%%%%%%%%

\bibliographystyle{asmems4}
\bibliography{dscc}

\appendix
\section{Situational Awareness Questionnaire}
\label{app:sitawareness}
These questions appeared in the middle of the task to measure the situational awareness of the operator: \\ \\
1. How many UAVs were you handling?\\
2. How much battery left of the No.1 (yellow/green/red/blue) UAV?\\
3. What was the altitude of the No.1 (yellow/green/red/blue) UAV?\\
4. What was the velocity of the No.1 (yellow/green/red/blue) UAV?\\
5. Where is the No.1/No.2/No.3/No.4 (yellow/red/green/blue) UAV?\\
6. Where was the last person you identified?\\
7. Where was the last vehicle you identified?\\
8. How many person did you identify?\\
9. How many person did you rescue?\\
10. How many people did you see but were unable to tag?\\
11. How many cars did you identify?\\
12. How many cars did you rescue?\\
13. How many cars did you see but were unable to tag?\\

\end{document}